\documentclass[letterpaper, 10 pt, conference]{ieeeconf}  % Comment this line out if you need a4paper

\IEEEoverridecommandlockouts                              % This command is only needed if 
\usepackage{multirow}
\usepackage{bmpsize}\usepackage[utf8]{inputenc}
\usepackage{graphicx}\usepackage[utf8]{inputenc}
\usepackage{amssymb,amsmath}
\usepackage{bm}
\usepackage[english]{babel}
\usepackage[ruled,vlined]{algorithm2e}
\usepackage{flushend}

\makeatletter
\renewcommand*\env@matrix[1][\arraystretch]{%
	\edef\arraystretch{#1}%
	\hskip -\arraycolsep
	\let\@ifnextchar\new@ifnextchar
	\array{*\c@MaxMatrixCols c}}
\makeatother

\usepackage{float}
\usepackage{amssymb}
\usepackage{soul}
\graphicspath{{Images/}}
\usepackage[colorlinks={true},linkcolor={black},citecolor={black}, urlcolor={black}]{hyperref}
\emergencystretch\maxdimen
\title{\LARGE \bf
Cartographer\_glass: 2D Graph SLAM Framework using LiDAR for Glass Environments
}

\author{Lasitha Weerakoon$^{*1}$, Gurtajbir Singh Herr$^{*1}$, Jasmine Blunt$^{2}$, Miao Yu$^{1}$ and Nikhil Chopra$^{1}$% <-this % stops a space
\thanks{\hspace*{-1.5em} $^*$The authors contributed equally toward this work. 
$^{1}$University of Maryland, College Park, MD 20742, USA, $^{2}$U.S. Department of Defense.\newline
{\tt\small \{lasitha,gurtaj89,mmyu,nchopra\}@umd.edu}}%
}

\begin{document}

\maketitle
\thispagestyle{empty}
\pagestyle{empty}

%%%%%%%%%%%%%%%%%%%%%%%%%%%%%%%%%%%%%%%%%%%%%%%%%%%%%%%%%%%%%%%%%%%%%%%%%%%%%%%%
\begin{abstract}
We study algorithms for detecting and including glass objects in an optimization-based Simultaneous Localization and Mapping (SLAM) algorithm in this work. When LiDAR data is the primary exteroceptive sensory input, glass objects are not correctly registered. This occurs as the incident light primarily passes through the glass objects or reflects away from the source, resulting in inaccurate range measurements for glass surfaces. Consequently, the localization and mapping performance is impacted, thereby rendering navigation in such environments unreliable. Optimization-based SLAM solutions, which are also referred to as Graph SLAM, are widely regarded as state of the art. In this paper, we utilize a simple and computationally inexpensive glass detection scheme for detecting glass objects and present the methodology to incorporate the identified objects into the occupancy grid maintained by such an algorithm (Google Cartographer). We develop both local (submap level) and global algorithms for achieving the objective mentioned above and compare the maps produced by our method with those produced by an existing algorithm that utilizes particle filter based SLAM. 
\end{abstract}

%%%%%%%%%%%%%%%%%%%%%%%%%%%%%%%%%%%%%%%%%%%%%%%%%%%%%%%%%%%%%%%%%%%%%%%%%%%%%%%%
\section{Introduction} \label{sec:intro}
The development of Simultaneous Localization and Mapping (SLAM) algorithms has enabled precise mapping of indoor and outdoor environments \cite{cadena2016past}. Improvements in sensing modalities such as LiDARs and cameras now enable high-quality 2D and 3D reconstruction of environments. LiDARs, in particular, have become increasingly popular due to their precision and long-range operation. Despite these attractive features, LiDARs perform poorly in the presence of transparent objects and reflective surfaces. In modern building designs, transparent walls and partitions are used more and more for aesthetic appeal and energy-efficient interior lighting. Therefore, detecting and including such transparent obstacles in the map is required for reliable and safe robot navigation. 
%This inability to account for glass in turn affects the localization and mapping accuracy of the SLAM algorithm and can lead to infeasible paths and collisions while navigating the environment.

A LiDAR measures the range to an object based on detecting the reflection of the incident laser beam from it. Diffuse surfaces disperse the beam equally in all directions.  Due to this dispersion, the intensity corresponding to the reflection that measures the distance to the object is usually low \cite{foster2013visagge}. However, when the beam is incident on a transparent object like glass, most of it passes through, and the remaining is reflected at an angle equal to the incident angle. This leads to LiDAR not reporting the object from most angles. But, when the incidence angle is near normal, most of the incident beam is reflected directly back to the source, and a sharp increase in intensity occurs. This intensity value is significantly higher than intensities values for reflection from diffused surfaces. 

% \subsection{Related Work}
There have been several approaches that have tried to address the challenge of detection and mapping glass points. Foster et al. \cite{foster2013visagge} consider that glass behaves like a diffuse object in a minimal range of angles (near-normal incidence) and thus gets treated as noise by the SLAM framework. They augment standard occupancy grid mapping by tracking the subset of angles from which glass objects are visible and reconstructing the objects from these angles assuming there is some localization error. In \cite{slam_glass_2017} a simple detection scheme is utilized by Wang and Wang by examining the intensity profile of the returned laser beam around normal incidence to the glass surface. This was then incorporated into GMapping, an existing particle filter based SLAM approach \cite{grisetti2007improved} naming the package as \textit{slam\_glass}. {Using this package, Shiina and Wang have developed an indoor navigation algorithm incorporating detected glass points \cite{shiina2017indoor}.}

The use of multi-echo lasers to detect glass objects has been described by Koch et al. \cite{koch2016detection, koch2017identification}. In recent work, Kim and Chung \cite{kim2018robust} utilize the reliability of LiDAR measurements to localize a robot in glass environments. {In \cite{awais2009improved}, Awais has proposed a probabilistic approach to model the glass surfaces using laser scans for improved laser-based navigation of mobile robots. Jiang et.al. have used neural networks to build glass confidence maps using laser range finders \cite{jiang2017glass, jiang2019robust}. However, they were not focusing on SLAM but only classification of wall surfaces as glass and non glass. Recently, Jung et.al. \cite{jung2020mobile} proposed a method to for transparent obstacle detection by analyzing the noise generated when a laser meets a transparent obstacle. In \cite{meng2020accurate}, Meng et.al. have used the cartographer maps generated from glass environments and manually labeled the glass walls for accurate LiDAR-based localization in glass-walled environments.}

Another common practice is to utilize additional sensors and employ sensor fusion to deal with the detection {\cite{huang2018glass}} and inclusion of glass into the map \cite{diosi2004advanced, wei2018fusing} {\cite{wei2018multi, yang2008dealing, yang2011onsolving}}. In these approaches, sonars are commonly used as the additional sensing modality. 

% \cite{huang2018glass} Glass detection and recognition based on the fusion of ultrasonic sensor and RGB-D sensor for the visually impaired. only for detection.

% \cite{yang2008dealing} Dealing with Laser Scanner Failure: Mirrors and Windows. sensor fusion for transparent and reflective surface detection and localization. sonar+laser scanner.SLAM and navigation . Accompanying the post-processing process, the range
% scan data can be revised such that the mirror images are
% eliminated. extended this in \cite{yang2011onsolving}.

% \cite{koch2017identification} Identification of transparent and specular reflective material in laser
% scans to discriminate affected measurements for faultless robotic
% SLAM. One drawback to laser scanners is their sensitivity
% to specular reflective and translucent surfaces, e.g., glass, mirrors,
% or shiny metal. When scanning such a surface, the laser beam can
% provide misleading measurements. Multi-echo laser scanner.  
% \cite{koch2017effective} Effective Distinction Of Transparent And Specular Reflective
% Objects In Point Clouds Of A Multi-Echo Laser Scanner

% \cite{awais2009improved} Improved laser-based navigation for mobile robots using the behavior of a laser scanner with respect to glass surface modeled using a probabilistic approach.

% \cite{wei2018multi} Multi-sensor Fusion Glass Detection for Robot Navigation and Mapping. 

% \textcolor{red}{Need more discussion about existing work...}

In this work,  we propose a 2D Graph SLAM framework using LIDAR for glass environments. Specifically, we exploit the sharp spike in intensity value to detect the presence of glass and other transparent surfaces as introduced in \cite{slam_glass_2017}. We incorporate these detected points into an existing Graph SLAM method known as the Google Cartographer \cite{cartographer_paper_2016}. Graph SLAM methods are now considered the state of the art methods in SLAM \cite{wilbers2019comparison}. Specifically, Google Cartographer has generally proven to provide robust and accurate maps over other SLAM methods \cite{krinkin2018evaluation}. Traditionally popular methods such as particle filter-based methods utilize a set of weighted hypotheses (particles) where each particle maintains the full representation of the environment being mapped \cite{grisetti2007improved}. Consequently, these methods can become computationally expensive when mapping large environments \cite{cartographer_paper_2016}. 

In the remainder of the paper, we will continue to refer to all the transparent \textit{glass-like objects} as glass. The main contributions of this work are the following:
\begin{itemize}
	\item Implementation of glass detection and mapping in an optimization-based SLAM algorithm (Google Cartographer),
	\item Improved accuracy over the maps generated by a GMapping based glass detection and mapping scheme,
	\item Extensive testing on glass environments using a mobile platform equipped with a LiDAR.
\end{itemize}

The rest of the paper is organized as follows. Section \ref{sub:background} briefly introduces the Google Cartographer, which is the Graph SLAM platform we utilize and the glass detection scheme. The proposed glass mapping schemes are discussed in Section \ref{sec:glass_detection_n_mapping}. Section \ref{sec:Experimental_results} presents the experimental results. Finally, in Section \ref{sec:conclusions} the conclusion and possible future directions are discussed. 

\section{Background}\label{sub:background}
\subsection{Google Cartographer}
The Google Cartographer \cite{cartographer_paper_2016} is a SLAM solution that achieves real-time mapping and loop closure using LiDAR data in 2D as well as 3D environments. In this work we only consider the 2D SLAM. The Google Cartographer consists of two subsystems: a local SLAM (front end) and a global SLAM (back end). The front end constructs local representations of the world known as submaps. The laser scans are inserted consecutively into the submap via a process called scan matching. Scan matching consists of optimizing the pose (position and orientation) of the laser scan for the submap by aligning the current scan with the submap. Even though the scan insertions are locally consistent, this process accumulates error. Submaps are the result of iteratively performing this alignment of the scan to submap coordinate frames. Representation wise, submaps are represented as a  probability grid where each grid cell is associated with a probability value. This value expresses the likelihood that the grid cell is occupied. A laser scan is inserted into this probability grid by determining the grid points' set where the laser ray ends. These grid points constitute the hits set. Any grid cell in which the ray intersects and is not in the hits set is included in the misses set. Previously unobserved grid points are assigned $p_{\text{hit}}$ or $p_{\text{miss}}$ which represent the hit and miss probabilities respectively. The grid probabilities of already observed cells are updated accordingly for hits ($p=p_{\text{hit}}$) or misses ($p=p_{\text{miss}}$) as,
\begin{align*}
    M_{\text{new}}(cell) = \text{odds}^{-1} \left(\text{odds}\left(M_{\text{old}}\left(cell\right)\right)\cdot \text{odds}\left(p\right)\right)
\end{align*}
where $M_{\text{old}}(cell)$ is the prior grid probability of the considered $cell$ and the odds for grid probability $p$ defined as $\text{odds}(p) = \frac{p}{1-p}$.

%\textcolor{blue} {Discussion of how exactly the probability grid updates the odds  is needed here} 

%The probability grid of a submap is maintained by uint16 \textit{Values} in the $[1, 32767]$ range which are converted from the probabilities. In normal operation an update marker of 32768 is added to these \textit{Values}. As the scans are received and points are hit or missed a Lookup table is used to assign these \textit{Values} to the cell indices of the grid.}

The back end's role is to check for loop closures and deliver the most consistent global solution. The error accumulated during scan insertion in submaps is minimized by running a pose optimization, where the poses of all scans and submaps are optimized. To check for loop closure, each scan is matched with already completed submaps. If a match is found, a constraint is inserted in the optimization problem that relates the relative poses. For more details, the reader is referred to \cite{cartographer_paper_2016}.

\subsection{Glass detection}
Only the intensity data of the reflected LiDAR beams is used to detect the transparent surfaces (glass) in this work, as introduced by the authors in \cite{slam_glass_2017}. As glass surfaces have highly polished surfaces, these specular reflect the laser beams. Therefore, the received rays' intensity will have the highest strength when the robot is receiving laser rays from the normal incident angle. Hence, by inspecting the laser scan data's intensity profile and identifying the intensity peak, the glass surfaces can be detected. For false proof detection, this algorithm uses an intensity threshold value (\textit{thresh}), an intensity gradient (\textit{grad}) and a profile width (\textit{width}) as tunable parameters. The method used for glass detection is shown in Algorithm \ref{alg:glass_detection}. Although this detection method allows only a small range of view angles, this has been a proven method of glass detection in recent literature \cite{foster2013visagge, slam_glass_2017}. 

\begin{algorithm}[h]
\SetAlgoLined
\textbf{Require:} Point Cloud with intensities and given \textit{thresh}, \textit{grad} and \textit{width}\\
\textbf{Initialization:} $isGlass(\text{Point Cloud}) = 0$ and int$_{\text{0}}$ = intensity of first data point

\For{all p in Point Cloud}{
int$_{\text{p}}$ = intensity of p \;
\eIf{int$_{\text{p}}$ $\geq$ \textit{thresh}  \& int$_{\text{p}}$-int$_{\text{p-1}}$ $\geq$ \textit{grad}}{
h $\gets$ p\;}
{
\If{int$_{\text{p}}$ $\geq$ \textit{thresh}  \& int$_{\text{p}}$-int$_{\text{p-1}}$ $\leq$ \textit{grad}}{
\If{p-h $\leq$ \textit{width}}{
r := floor$(\frac{p+h}{2})$\;
$isGlass(r)$ $\gets$ 1} ;
}}
}
\textbf{return} $isGlass(\text{Point Cloud})$.
\caption{Glass Detection}
\label{alg:glass_detection}
\end{algorithm}
\vspace{-10pt}

\section{Glass mapping in 2D Graph SLAM} \label{sec:glass_detection_n_mapping}
We will discuss the methodology of incorporating the detected glass points to a 2D Graph SLAM framework, the Google Cartographer.

The main objective of the glass mapping scheme is to add the detected glass points robustly into the map. Ideally, this could be done by identifying the cell index of the detected glass point in the probability grid and commanding to increase the grid probability of that particular cell to the maximum allowable probability. However, a glass point is visible and detected as a \textit{hit point} only when the incident laser scan ray is normal to the glass surface. Therefore, when the laser rays pass through glass or specular reflect in a direction away from the LiDAR, that particular point would be considered a \textit{miss point} as opposed to a \textit{hit point}. Hence, even if the detected point's grid probability is set to a high value at the time of detection, the Cartographer algorithm will reduce its probability afterward when the same glass point is registered as a \textit{miss point}. This is due to the inherent nature of the Cartographer algorithm as it updates the probability grid by considering the hits and the misses, as explained in Section \ref{sub:background}.

Therefore, to incorporate the detected points in the map resolutely, we propose the \textbf{Cartographer\_glass} framework which uses two sub-schemes. 1). The \textit{local glass mapping scheme} maps the detected glass points onto the current submap, and 2). the \textit{global glass scheme} retains the detected glass points in the global map.   

\subsubsection{Local glass mapping}
In this scheme, the cell index of the detected glass point in the current submap's probability grid is first identified. Next, the grid probability of the identified cell is artificially set to the maximum probability value. Consequently, the detected glass point is now added to the local submap as a hit point with the maximum grid probability, encoding an identification of the observed detected glass point, which will be useful in identifying subsequent occurrences of the same point. If an already observed glass point is observed again, as a hit or as a miss, we will not update its occupancy probability from the already set maximum value. This procedure ensures that the detected and mapped glass points on the submap do not fade away. Further, the global coordinates of the detected glass points with respect to the initial global coordinate frame is saved in a global \textit{Glass\_points} array by transforming the coordinates given in the current global coordinate frame using a homogeneous transformation matrix $H_k$. Here $H_k$ is the homogeneous transformation from the initial global coordinate frame to the $k^{th}$ (current) global coordinate frame, which is computed while running the loop closure optimization (back end of Cartographer). 
The \textit{local glass mapping} scheme is shown in Algorithm \ref{alg:local_glass} which will return the probability grid for the $k^{th}$ submap.

\begin{algorithm}[h]
\SetAlgoLined
\textbf{Require: } Range data set including the detected glass points with an identifier, global \textit{Glass\_points} array and the homogeneous transformation matrix $H_k$ .\\
\textbf{Initialization: } $k^{th}$ submap with probability grid 

\While{ $!$ Finish\_Submap }{
\For{all $x_i \in$ Range Data }{
cell $\gets$ index($x_i$)\;
 \If{$x_i =$ Detected glass point}{
  $M_{\text{new}}(cell) \gets {max}_{{p}}$ (with identifier)\;
  \textit{Glass\_points} $\gets$ $[\textit{Glass\_points};H_k x_i]$;
  }
  \If{$M_{\text{new}}(cell) != {max}_{{p}}$ (with identifier)}{
    $M_{\text{new}}(cell) \gets$ from normal calculation \;}
   }
   }
   \textbf{return } probability grid.
 \caption{Local Glass Mapping Scheme}
 \label{alg:local_glass}
\end{algorithm}

\subsubsection{Global glass mapping}

The \textit{local glass mapping} scheme is only applicable in the submap level as the computed probability grid is re-initialized at the start of each submap. Therefore, the previously observed glass points' identifiers in earlier submaps no longer exist in the newly initialized probability grid. Thus, if a previously observed glass point is now registered as a miss, it will result in fading away of the already added detected glass points in a previous submap according to the probability update using $p_{\text{miss}}$. Therefore, it is essential to add the saved points from the global \textit{Glass\_points} array to each subsequent submap to retain the detected surfaces in the global map. However, these points need to be transformed to the current global coordinate frame before adding them. {{Next, the grid probability of these added points will be set to maximum probability (with the encoded identifier)}}. This will enable the \textit{local glass mapping} scheme to identify the newly added points as glass points when evolving the submap. Algorithm \ref{alg:global_glass} shows the \textit{global glass mapping} scheme. 

\begin{algorithm}[h]
\SetAlgoLined
\textbf{Require: } Global \textit{Glass\_points} array and the homogeneous transformation matrix $H_k$ .\\
\textbf{Initialization: } new $k^{th}$ submap

\For{$p_i \in {Glass\_points} \cap span(submap)$}{
cell $\gets$ index($H_k^{-1} p_i$)\;
   $M_{\text{new}}(cell) \gets {max}_{{p}}$ (with identifier)\;}
Local Glass Mapping Scheme\;
   \textbf{return } probability grid.
 \caption{Global Glass Mapping Scheme}
 \label{alg:global_glass}
\end{algorithm}

However, one significant challenge for implementing the global glass scheme is estimating the homogeneous transformations of the detected glass points accurately. Moreover, as the loop closures are happening for sections of the map after observing multiple times, the current submap might span to regions that are yet to be optimized by adding loop closure constraints. In these situations, the added detected points might have a drift corresponding to the error in the robot's actual and estimated trajectories/poses. This is more evident in scenarios which large spaces are mapped, because the possibility of introducing drifts is greater in such scenarios. Therefore, in such circumstances, the above \textit{global glass mapping} scheme is ineffective. However, as an alternative method to overcome this challenge is only to use the \textit{local glass mapping} algorithm and increase the $p_{\text{miss}}$ value so that the past points does not fade as frequently. We will refer to this method which only the \textit{local mapping scheme} is used with an increased $p_{\text{miss}}$ as the \textbf{Cartographer\_glass\_lite}.

\section{Experimental results} \label{sec:Experimental_results}
The experiments were conducted using a commercially available MIT RACECAR robotic platform equipped with a Hokuyo UST-10LX Scanning Laser Rangefinder (LiDAR). The SLAM algorithms were run offline on a PC using the collected LiDAR data. First we illustrate the effectiveness of the proposed SLAM algorithm in a laboratory setting with different transparent panels outlining the calibration procedure. Next, the generated maps using the proposed glass detection and mapping scheme for the data collected from actual office buildings will be presented.

% First we present the calibration procedure of the parameters for glass detection. Then we illustrate the effectiveness of the proposed SLAM algorithm in a laboratory setting with different transparent panels. Next, the generated maps using the proposed glass detection and mapping scheme for the data collected from actual office buildings will be presented.

% Although identifiers are encoded in the submap level probability grid, the proposed algorithm does not label the detected glass points in the global map. \textcolor{red}{reason???} 
As these maps' motivation was to provide a cost map to a navigation stack, we do not explicitly label each point as glass or not glass but treat all the points the same. Therefore, all the maps are converted and saved as an image file that encodes the occupancy data using the \textit{map\_server} ROS Node by coloring each pixel correspondingly to the occupancy state of the cells of the map \cite{ros.org}. Free cells are colored white, and occupied cells are colored black whereas unknown cells are colored in gray.

\subsection{Detection and mapping of different transparent panels}
For this experiment, two clear acrylic panels, a glass panel, and a clear poly-carbonate panel were placed in a laboratory setting. The robot was then run to collect the LiDAR data. 

\begin{figure}[t]
	\centering
		\includegraphics[width=0.45\textwidth]{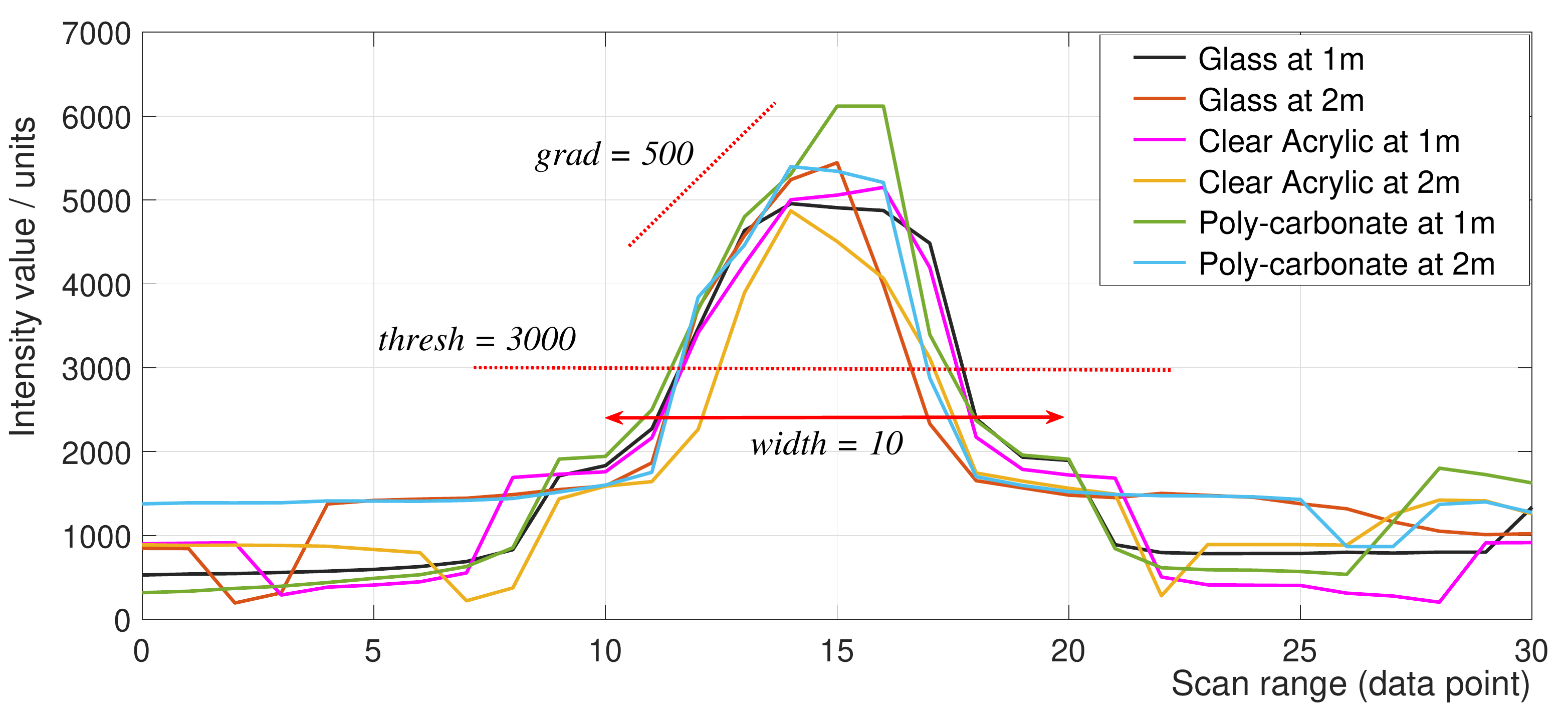}
	%	\captionsetup{justification=centering}
	\caption{Intensity profiles from different materials and the selected parameters}
	\label{Fig:intestity plot}
\end{figure}

Before running SLAM algorithms, experiments were performed to collect data from the different transparent panels to observe their intensity profiles and select the intensity threshold value for transparent surface detection. For each of the materials, the panel was placed at a distance of 1m perpendicular to the direction in which the robot was pointing to ensure near-normal incidence, and the LiDAR data was recorded. This was repeated from a 2m distance as well. Parameter values were selected such that it would enable the detection of all material types used. Thus an intensity threshold of  $thresh = 3000$, an intensity gradient of $grad = 500$ and a profile width of $width = 10$ were selected. The materials used for data collection are shown in Table \ref{tab:material}, and the intensity profiles near the normal incident angle are shown in Fig. \ref{Fig:intestity plot} along with the selected parameter values.

\begin{table}[h]
    \caption{Different material used for collecting data}
    \centering
    \begin{tabular}{|c|c|c|c|}
        \hline
         Material & Thickness (in) & Height (in)  & Width (in)   \\
         \hline
        Glass sheet & 0.125 & 12 & 18 \\
        Clear Acrylic sheet & 0.093 & 12 & 24 \\
        Poly-carbonate sheet & 0.5 & 14 & 24\\
        \hline
    \end{tabular}
        \label{tab:material}
\end{table}

\begin{figure}[t]
	\centering
	\begin{tabular}{c c}
		\includegraphics[height=1.4in]{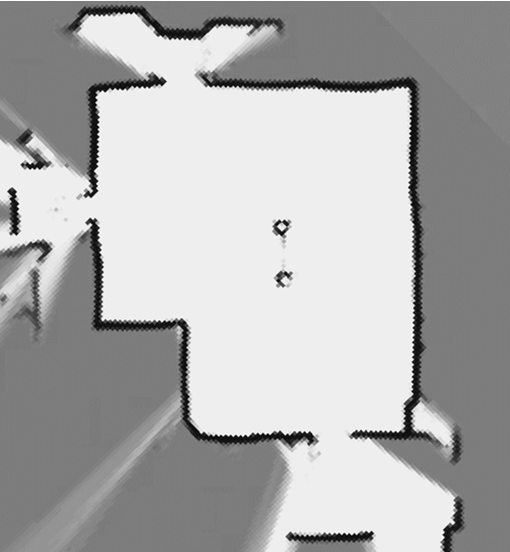} & \includegraphics[height=1.4in]{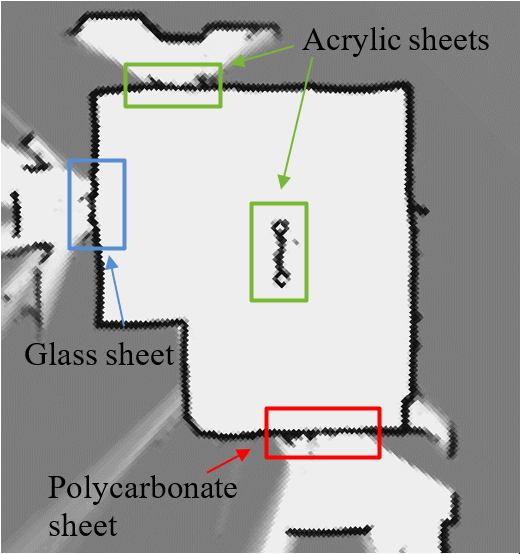}\\
		\footnotesize{(a) Default Cartographer map} & \footnotesize{(b) \textit{Cartographer\_glass} map}
	\end{tabular}
	%	\captionsetup{justification=centering}
	\caption{Generated maps without and with glass detection using Google Cartographer inside a laboratory setting with different transparent panels}
	\label{Fig:RRL_map}\vspace{5pt}
\end{figure}

The map generated from the default Cartographer without the glass detection is shown in Fig.\ref{Fig:RRL_map} (a) in which the surfaces pertaining to the placed panels are marked as free. From Fig.\ref{Fig:RRL_map} (b) it is clearly seen that the Cartographer\_glass algorithm has detected and added the transparent surfaces to the final map.

\subsection{Experimentation in office buildings}
In this section, the generated maps from data collected by running the robots in real-world office buildings are presented. We consider three scenarios- the data set provided by the authors in \cite{slam_glass_2017} and two data sets collected by the RACECAR robotic platform. 
% All the maps are converted and saved as an image file that encodes the occupancy data using the \textit{map\_server} ROS Node by coloring each pixel correspondingly to the occupancy state of the cells of the map \cite{ros.org}. Free cells are colored white and occupied cells are colored black where as unknown cells are colored in gray. 

\subsubsection{Dataset from slam\_glass paper}
\begin{figure}[t]
	\centering
	\begin{tabular}{c c c}
			\includegraphics[width=1.3in]{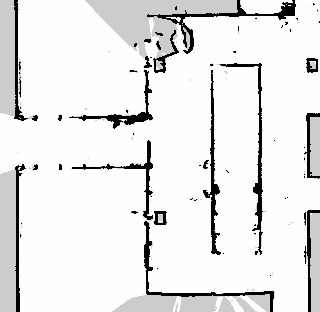} & \vline & \includegraphics[width=1.3in]{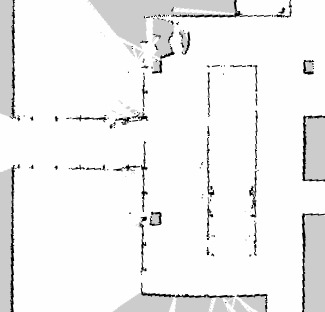}\\
		\footnotesize{(a) \textit{Cartographer\_glass} map } & &
		\footnotesize{(b) \textit{slam\_glass} map }
	\end{tabular}
	%	\captionsetup{justification=centering}
	\caption{Generated maps for a part of the dataset from \cite{slam_glass_2017}}
	\label{Fig:slam_glass_paper}
\end{figure}

Here we present the maps generated from the data set provided in \cite{slam_glass_2017}. We used the recommended parameter values of $thresh = 8000$, $grad = 4000$ and $width = 5$.  Fig.\ref{Fig:slam_glass_paper} shows the maps generated by the proposed Cartographer\_glass algorithm (a) and the \textit{slam\_glass} map (b). By visual inspection it is clearly seen that the glass detection and mapping is more robust from the proposed Cartographer based implementation.

\subsubsection{Data collected by RACECAR platform}
We present the maps generated from the MIT RACECAR platform's data running inside two buildings with glass environments. Since the glass walls in these buildings were different from those used in the laboratory setting experiment, a calibration process, as discussed earlier, was repeated. The robot was placed in front of a glass wall to collect the LiDAR data, and by inspecting the intensity profile, it was determined to use the parameter values as $thresh = 1300$, $grad = 100$ and $width = 10$ for both the buildings. Furthermore, since these are large environments we have used the Cartographer\_glass\_lite framework with an increased $p_{\text{miss}} = 0.499$.

\begin{figure}[t]
	\centering
		\includegraphics[width=0.48\textwidth]{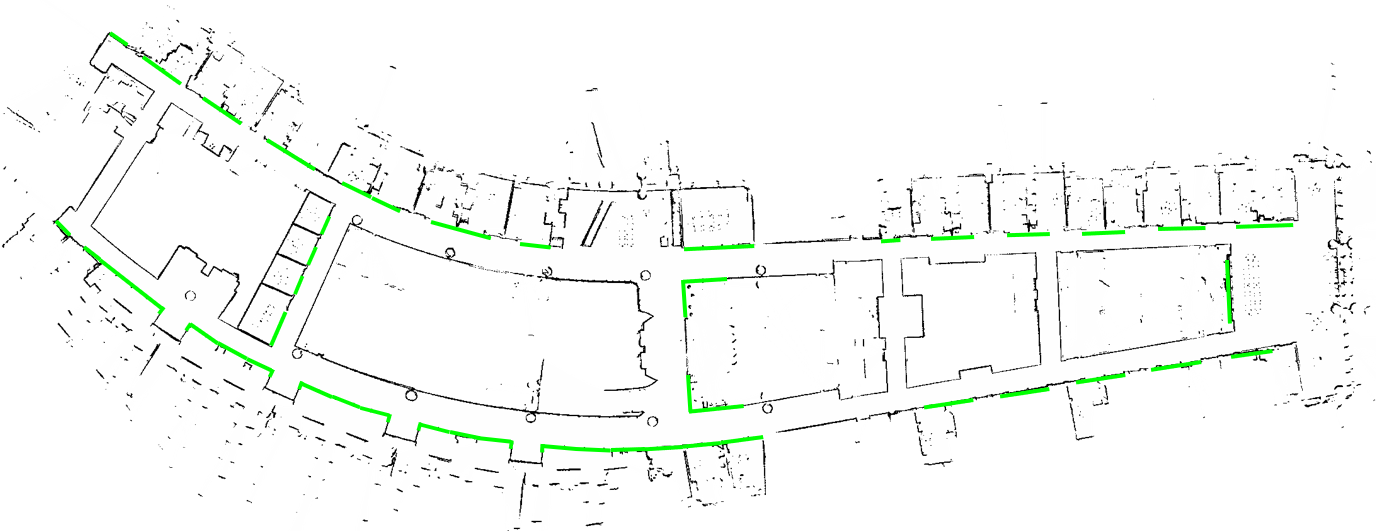}
	%	\captionsetup{justification=centering}
	\caption{Cartographer map with standard setting from \textit{Building 1} with the glass walls manually marked in green}
	\label{Fig:iribe_carto_map_overlayed}
\end{figure} 

\begin{figure}[h!]
	\centering
		\includegraphics[width=0.48\textwidth]{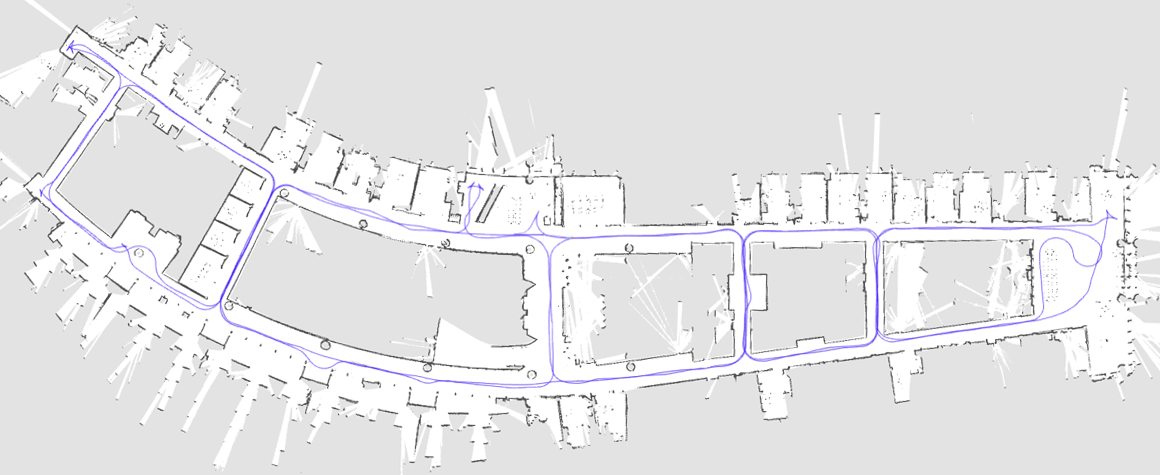}
	%	\captionsetup{justification=centering}
	\caption{Default Cartographer map from \textit{Building 1} with the robot path visualized in blue}
	\label{Fig:raw_carto_iribe}
\end{figure} 

\begin{figure}[h!]
	\centering
		\includegraphics[width=0.48\textwidth]{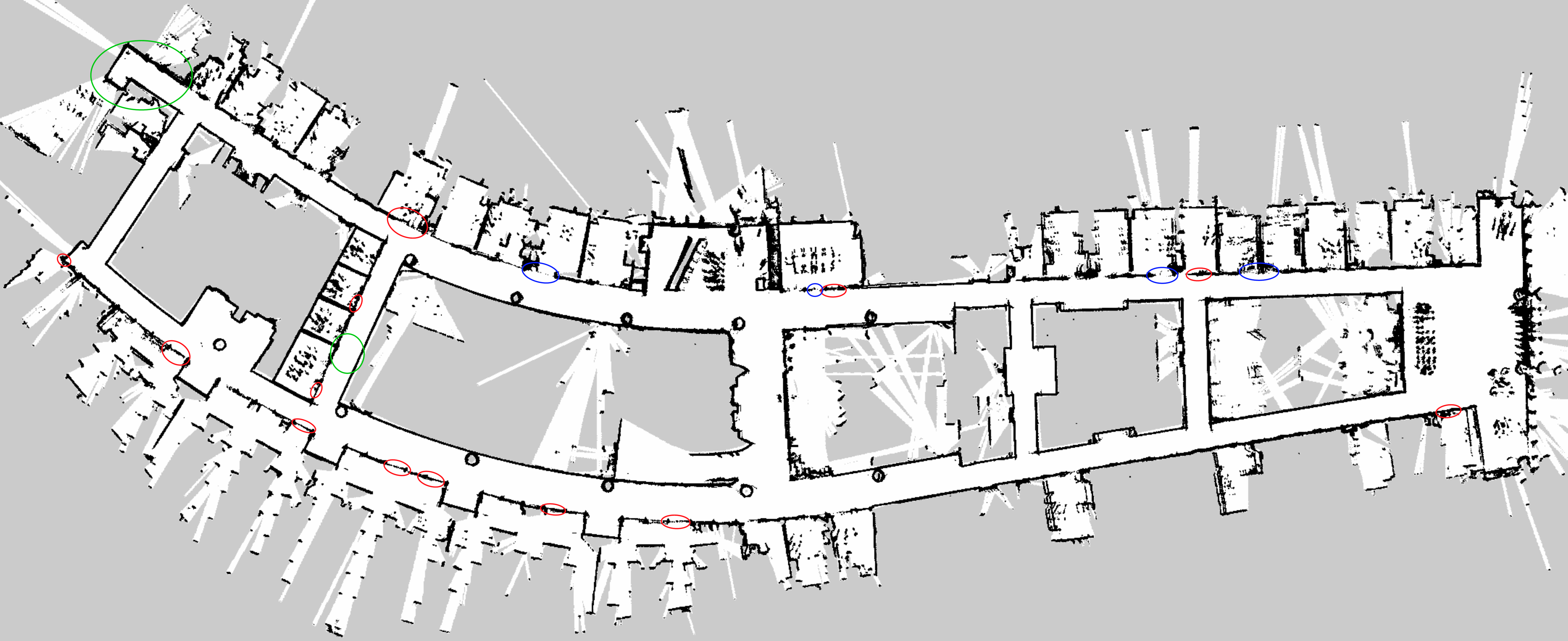}\\
		\footnotesize{(a) \textit{Cartographer\_glass\_lite} map }	\vspace{10pt}\\
		\includegraphics[width=0.48\textwidth]{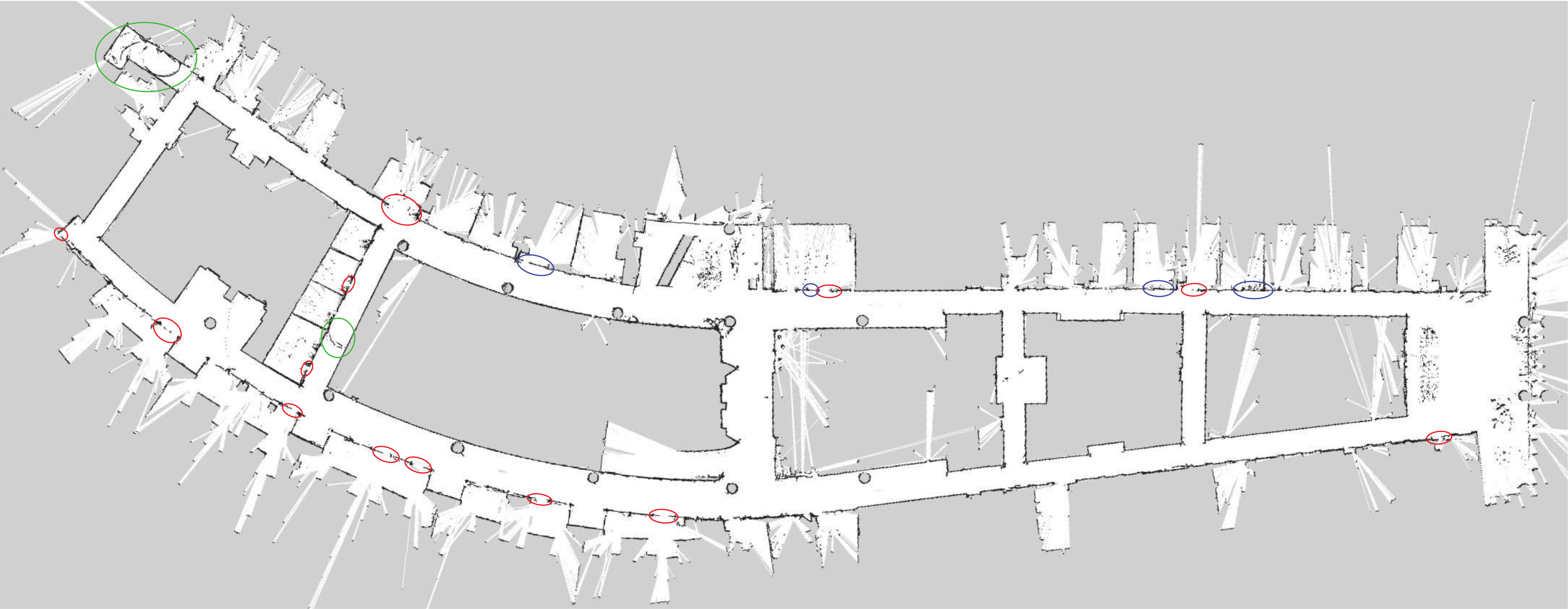}\\
		%\vspace{-10pt}\\
		\footnotesize{(b) \textit{slam\_glass} map } 
	%	\captionsetup{justification=centering}
	\caption{Glass detection and mapping enabled maps generated from the data collected from \textit{Building 1}. The spots which the proposed Cartographer based SLAM outperformed the \textit{slam\_glass} map are highlighted in red color where-as the spots which the \textit{slam\_glass} map had done better are highlighted in blue. The green marks indicate false points added in the \textit{slam\_glass} map.}
	\label{Fig:iribe_map}
\end{figure}

%Iribe center
First, the maps generated from the \textit{Building 1} are presented. The ground truth for glass walls' location is manually marked in Fig.\ref{Fig:iribe_carto_map_overlayed} for illustration. The robot passed all the long hallways twice while the connecting short corridors were traversed twice or four times, as indicated in Fig.\ref{Fig:raw_carto_iribe}, which is a Cartographer map generated with default settings.  Fig.\ref{Fig:iribe_map} shows the maps generated by the proposed \textit{Cartographer\_glass\_lite} framework and from the \textit{slam\_glass} algorithm. We have highlighted the map parts in which the proposed Cartographer algorithm outperformed the \textit{slam\_glass} map as well as the areas which the \textit{slam\_glass} did better glass detection and mapping.
% Figure \ref{Fig:iribe_carto_map_overlayed} illustrates the Cartogrpaher\_glass map overlayed with the actual floor plan of the building.

% Here we presented the generated maps collected by the MIT RACECAR platform running in the third floor of the Brendan Iribe Center and in the ground floor of A. James Clark Hall.

\begin{figure}[t]
	\centering
		\includegraphics[width=0.47\textwidth]{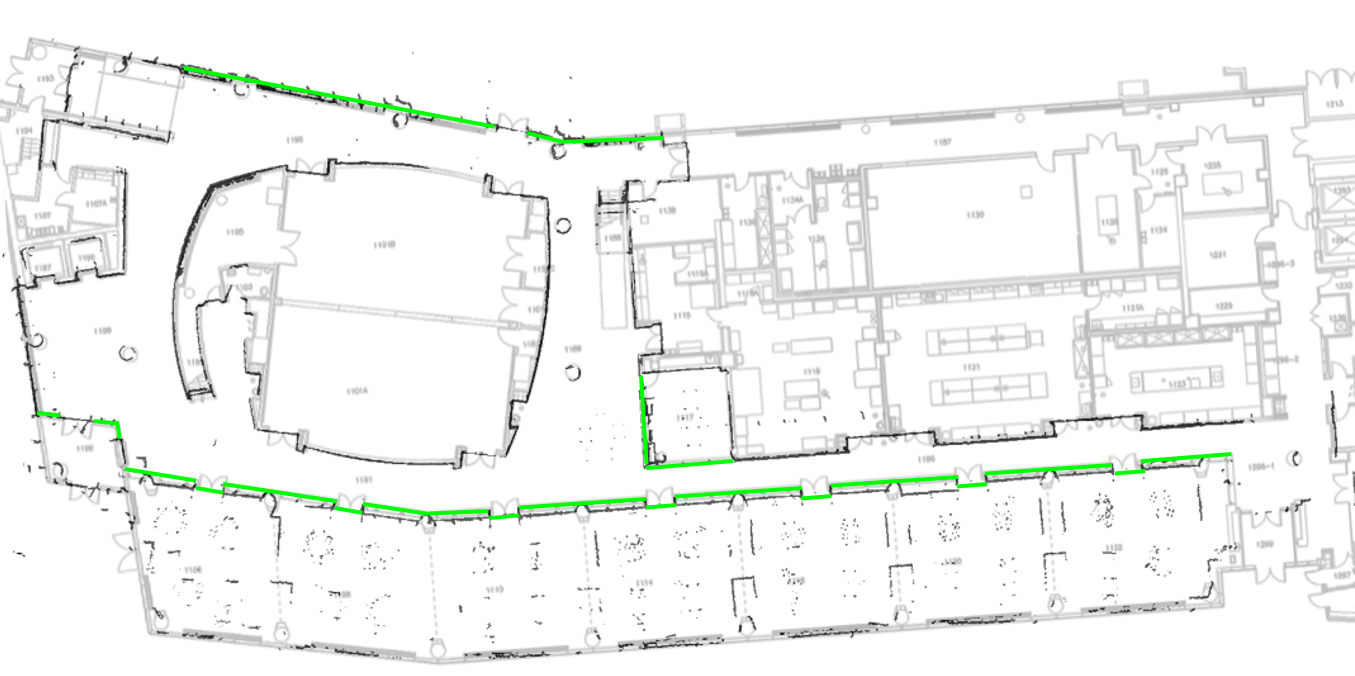}
	%	\captionsetup{justification=centering}
	\caption{Cartographer map with standard setting from \textit{Building 2} overlayed with the floor plan with glass walls marked manually in green}
	\label{Fig:clark_overlayed}
\end{figure}

\begin{figure}[h!]
	\centering
		\includegraphics[width=0.48\textwidth]{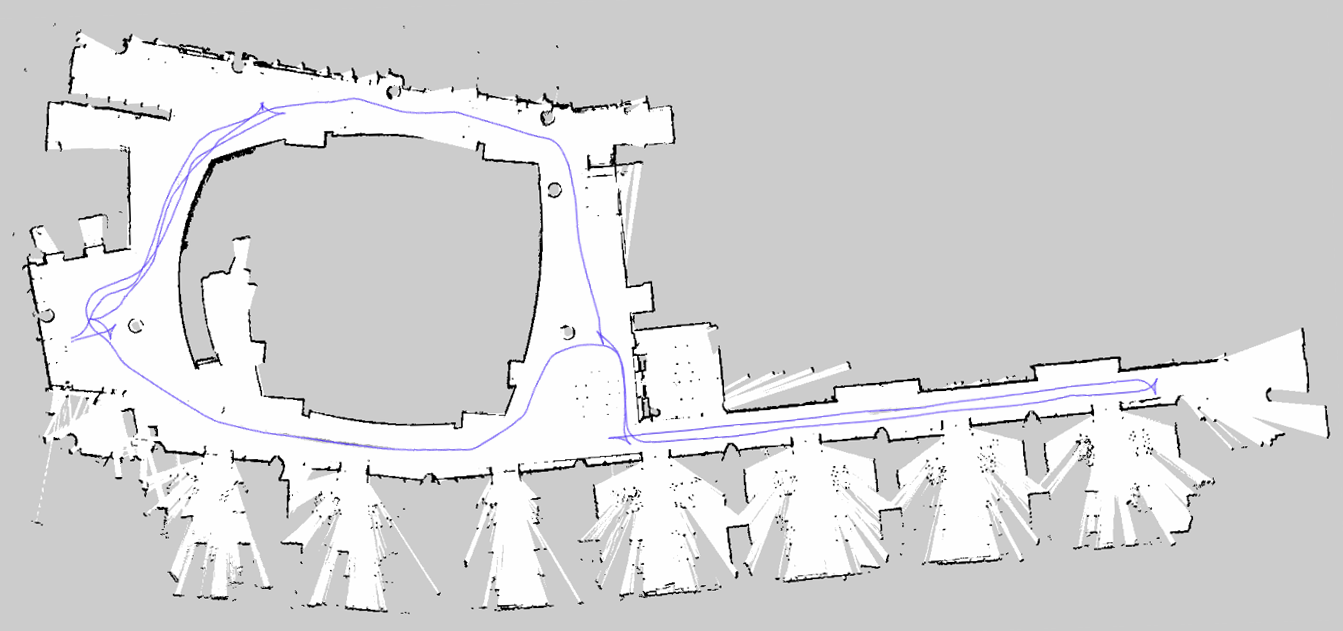}
	%	\captionsetup{justification=centering}
	\caption{Cartographer map with standard setting for \textit{Building 2} with the robot path visualized in blue}
	\label{Fig:clark_robotpath}
\end{figure}

\begin{figure}[h!]
	\centering
	\includegraphics[width=0.48\textwidth]{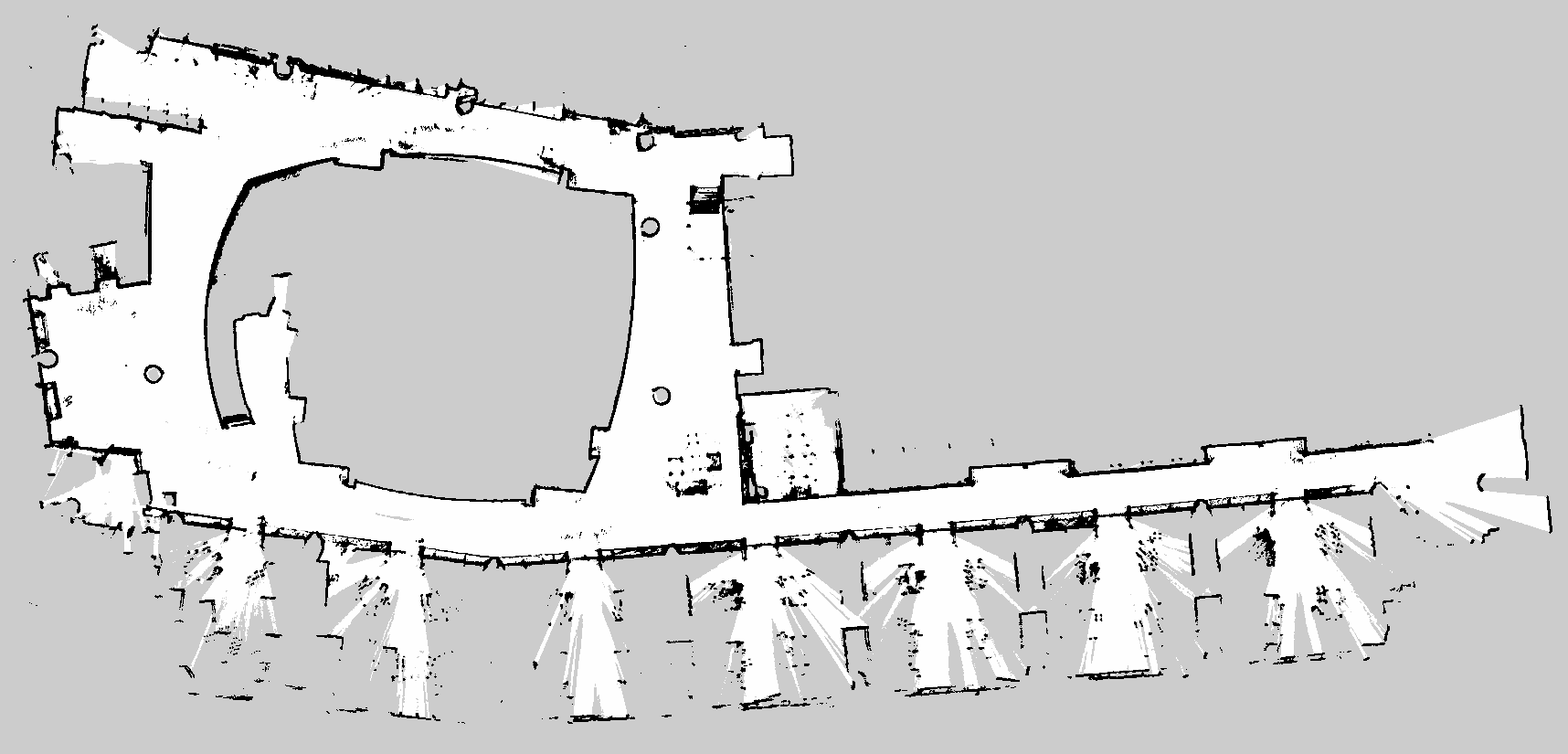}\\
		\footnotesize{(a) \textit{Cartographer\_glass\_lite} map } \vspace{10pt}\\
		\includegraphics[width=0.48\textwidth]{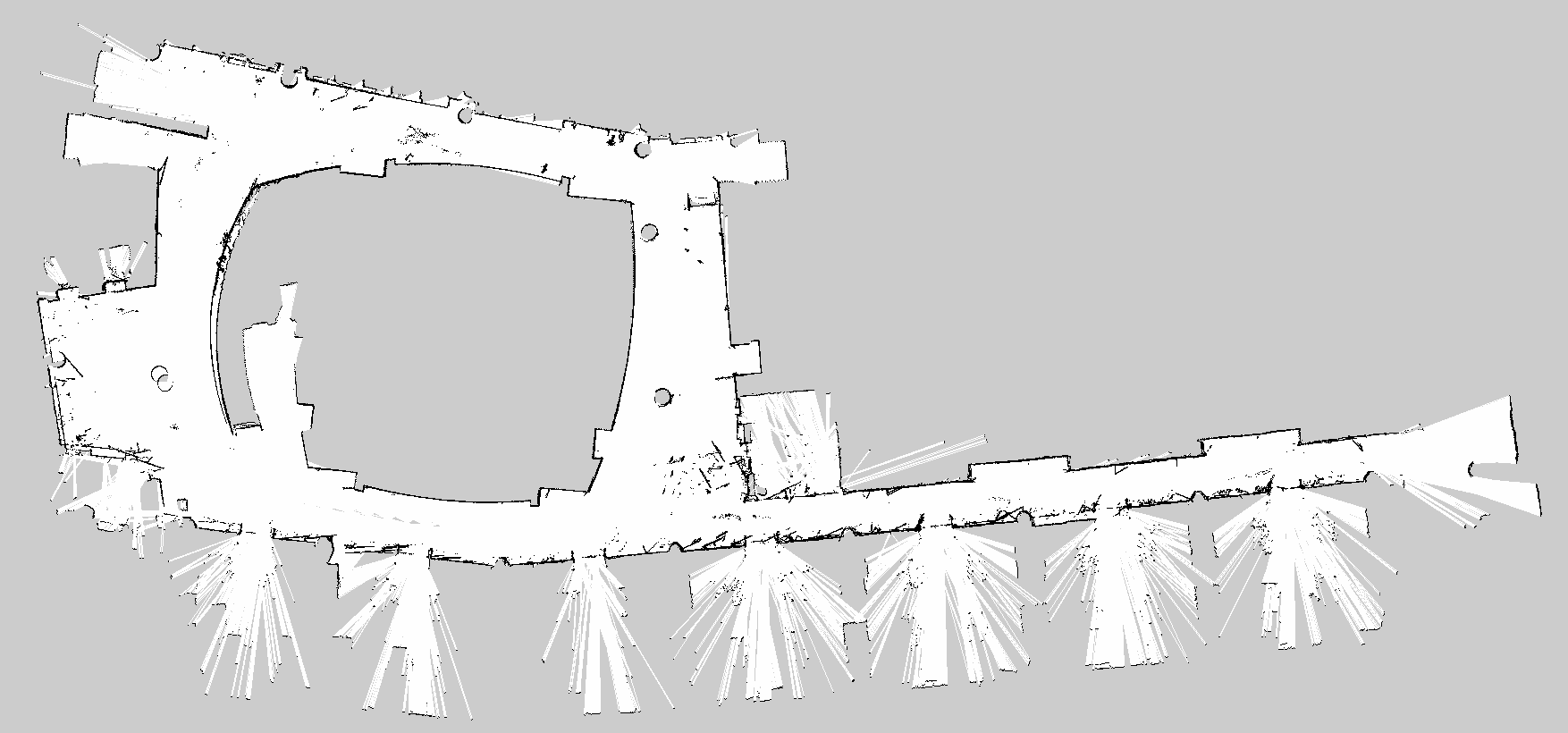}\\
		\footnotesize{(b) \textit{slam\_glass} map }
	%	\captionsetup{justification=centering}
	\caption{Generated maps from the proposed \textit{Cartographer\_glass\_lite} and from the \textit{slam\_glass} algorithm for data collected from \textit{Building 2} }
	\label{Fig:Clark_map}
\end{figure}

%Clark hall
Next, we present the maps generated from the MIT RACECAR platform's data running in \textit{Building 2}. Fig.\ref{Fig:clark_overlayed} illustrates the Cartographer map overlayed with the actual floor plan of \textit{Building 2} with the locations of the glass manually marked for illustration. Fig.\ref{Fig:clark_robotpath} shows the trajectory of the robot on a map generated by the Cartographer without glass detection (standard settings). The maps generated by the proposed \textit{Cartographer\_glass\_lite} framework and from the \textit{slam\_glass} algorithm are shown in Fig.\ref{Fig:Clark_map}.

It can be observed that the map generated by the proposed cartographer improvement (\textit{Cartographer\_glass\_lite}) is more accurate than that generated by the Gmapping based (\textit{slam\_glass}) map. {{While the glass walls are not accurately mapped in the slam\_glass SLAM map, it has also exposed a downside of using Gmapping which does not have the real-time loop closure property when constructing the map. This is clearly evident in the left side of the map where the map shows a misalignment.}}

\subsection{Discussion and Evaluation of the results}
From visual inspection, we observe that the maps generated by the proposed SLAM solution based on Google Cartographer are more accurate than the maps generated by the equivalent GMapping based \textit{slam\_glass} algorithm. However, we observe some undetected parts in the proposed SLAM maps. These might be because the robot might not have had a normal incidence from those areas for several reasons. The glass panels might not have been precisely vertical, or the ground might have had a slight inclination. {{Another reason could be that since we are using the alternative \textit{global glass mapping} method (\textit{Cartographer\_glass\_lite}), the number of subsequent \textit{misses} of the detected point in subsequent submaps might have been significant that, the cartographer algorithm has faded the point away by reducing the occupancy probability.}}

To evaluate the results, we manually identified the lengths of glass walls that were detected by the proposed algorithm and the \textit{slam\_glass} algorithm for comparison. Since the ground truth of the data set provided in \cite{slam_glass_2017} is unknown, we have not included it in this evaluation. Table \ref{tab:glass detection} shows the performance of the proposed glass detection and mapping algorithm, in which Carto indicates \textit{Cartogrpaher\_glass\_lite} and Gmap indicates \textit{slam\_glass}.  It is clearly seen that the proposed Graph SLAM based glass detection and mapping framework generates more accurate maps compared to the existing method.

\begin{table}[h!]
    \caption{Performance comparison of glass mapping}
    \centering
    \begin{tabular}{|c|c|c|c|c|c|}
        \hline
         \multirow{2}{1.2cm}{Building} & \multirow{2}{1.2cm}{Glass length (m)} & \multicolumn{2}{|p{2.15cm}|}{Detected glass (m)}  & \multicolumn{2}{|p{2.2cm}|}{Glass accuracy (\%)}  \\
         \cline{3-6} & & Carto & Gmap & Carto & Gmap\\
         \hline
       No.1 & 140.5 & 120.3 & 118.2 & 85.9 & 84.0 \\
       No.2  & 110.0 & 107.2 & 49.1 & 97.5 & 44.5 \\
        \hline
    \end{tabular}
        \label{tab:glass detection}
\end{table}

\section{Conclusion and future work} \label{sec:conclusions}

We have introduced a framework to incorporate glass detection and mapping capabilities to an optimization-based SLAM framework, namely the Google Cartographer. We name the proposed algorithm as \textit{Cartographer\_glass}. We observe that the proposed algorithm's maps had better accuracy than those generated by the \textit{slam\_glass} map, which uses the same glass detection scheme but uses a particle filter based SLAM. Identifying heterogeneous material surfaces and labeling them accordingly in the generated map will be considered in our subsequent work. Further, we note that the determination of accurate coordinates of the detected glass points in the global map, an integral part of the \textit{global glass scheme}, is still an unsolved problem.  Our focus in the future will be to face this challenge to improve on the proposed SLAM algorithm.

% {\textcolor{red}{However, as the main focus of the developed maps were to enable a navigation algorithm to pick up a cost map of the world, in the proposed SLAM solution, we have not implemented a method to distinguish between different material, but treat all the obstacles the same.}} 

\section{Acknowledgements} \label{sec:acknowledgements}
This research was supported in part by the U.S. Department of Defense. The authors would like to thank the Robotics Realization Lab of the Maryland Robotic Center and the lab manager Dr. Ivan Pensky for their helping in conducting the physical experiments.

\bibliographystyle{IEEEtran}
\bibliography{cartographer_glass}

\end{document}